Method to Detect Eye Position Noise from Video-Oculography

when Detection of Pupil or Corneal Reflection Position Fails

Evgeny Abdulin, Lee Friedman and Oleg V. Komogortsev

Department of Computer Science

Texas State University

San Marcos, Texas

Corresponding Author:

Evgeny Abdulin

Dr. Komogortsev's Lab (Comal 307D)

Comal Building

Department of Computer Science

Texas State University

601 University Dr.

San Marcos, Texas, 78666

USA

Email: abdulin@txstate.edu

**Abstract:**

We present software to detect noise in eye position signals from video-based eye-tracking systems that depend on accurate pupil and corneal reflection position estimation. When such systems transiently fail to properly detect the pupil or the corneal reflection due to occlusion from eyelids, eye lashes or various shadows, the estimated gaze position is false. This produces an artifactual signal in the position trace that is rapidly, irregularly oscillating between true and false gaze positions. We refer to this noise as RIONEPS (Rapid Irregularly Oscillating Noise of the Eye Position Signal). Our method for detecting these periods automatically is based on an estimate of the relative inefficiency of the eye position signal. We look for RIONEPS in the horizontal and vertical traces separately, and although we typically use it offline, it is suitable to adaptation for real time use. This method requires a threshold to be set, and although we provide some guidance, thresholds will have to be estimated empirically.

**Introduction:**

Video-based eye tracking systems that depend on accurate pupil and corneal reflection position estimates (video-oculography within this paper, or VOG) are very common and widely used [1, 2].  Generally speaking, these devices have high performance in detecting eye movements, with high accuracy and temporal precision [3].  However, as described by several authors, under certain conditions the detection of pupil position or corneal reflection can fail [2-6].  The most detailed description of the problems that can occur is in [2].  The pupil may be occluded by the upper eyelid, the eye-lashes (perhaps downward directed) and shadows. These problems are exacerbated by large pupils and downward facing eye lashes.  The corneal reflection may also be occluded, or, if there is excessive tear fluid in the eye (e.g. due to allergies, "Wet Eyes"[2]), or the subject is wearing spectacles or contact lenses, one or more bright reflections can appear and be falsely recognized as corneal reflections.

Typically, these detection failures occur transiently, with a rapid but irregular switching between good detection and failure, and this gives the eye movement recording a rapid, irregularly oscillating noise in the eye position trace, a noise we refer to as RIONEPS.   Although software and hardware solutions to improve pupil detection have been proposed [3-6], many researchers do not have access to such corrected systems.  Therefore, algorithms that can detect the presence of this noise in eye position signals would be useful. We present such an algorithm in this report.

**Examples of RIONEPS in Several Eye-Tracking Devices:**

We begin by illustrating several examples of RIONEPS from various VOG systems. When available, we show data representing pupil size estimates along with the eye position traces. The first two examples are from the EyeLink-1000 [7].

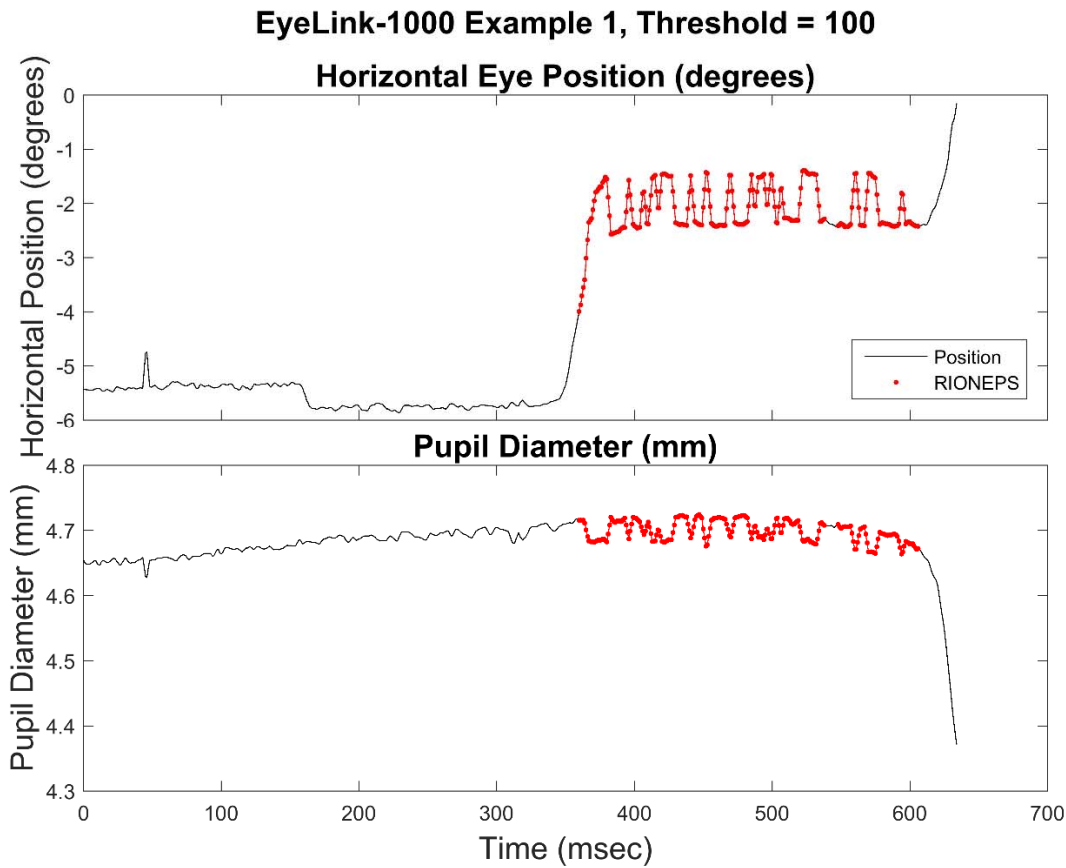

*Figure 1. RIONEPS in the horizontal position trace along with instantaneous measures of pupil diameter. Note the rapid, irregularly oscillating character of the noise in the eye position signal in synchrony with oscillations in pupil diameter estimates.*

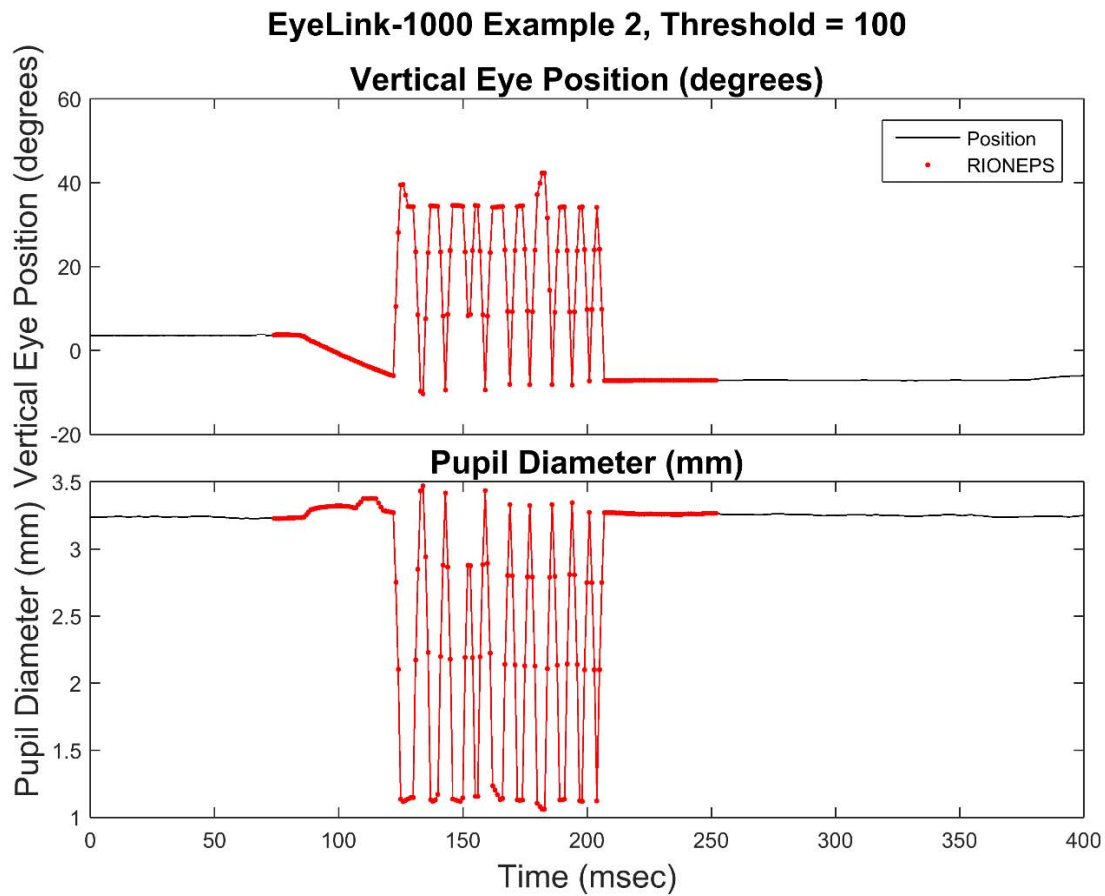

*Figure 2: A different subject shows marked RIONEPS in the vertical position trace in synchrony with marked changes the estimation of pupil diameter. The noise is more regular looking than the first example, but is still somewhat irregular.*

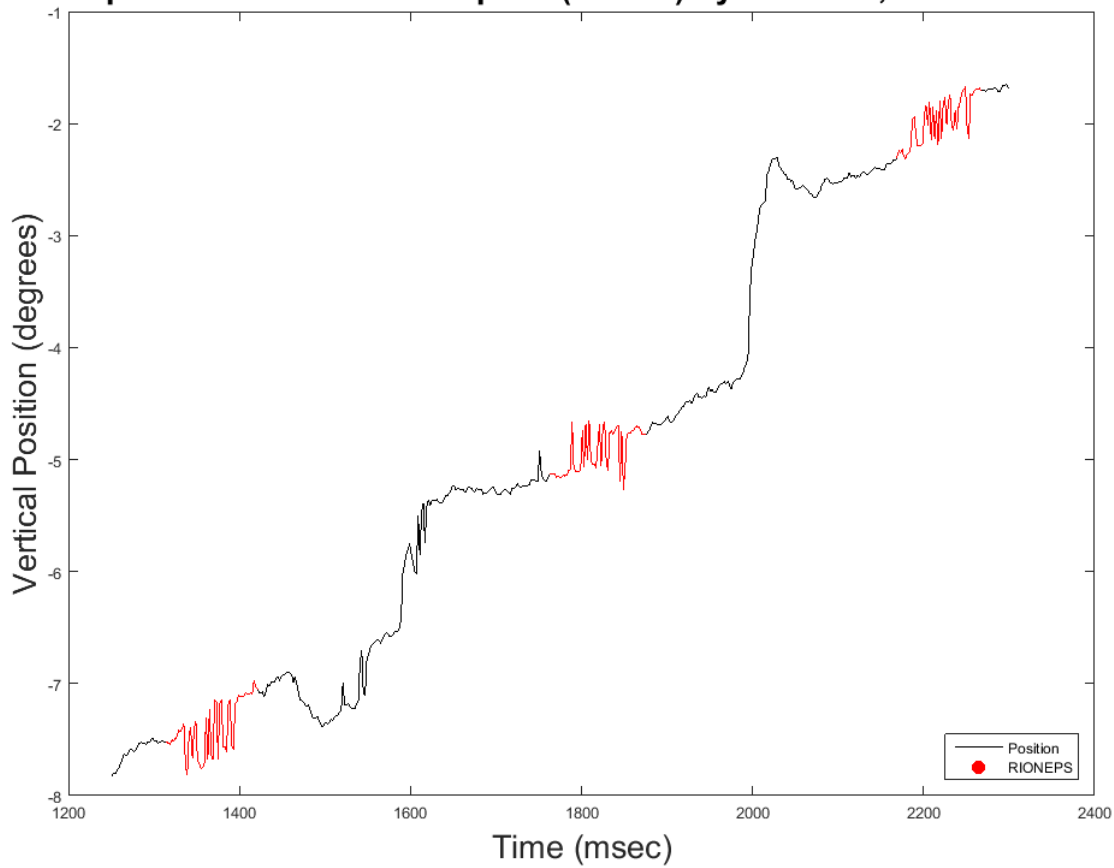

*Figure 3: The first of two examples from the same subject. The eye-tracker is described at [8]. These data were downloaded from Dr. Nyström's web site at Lund University (http://www.humlab.lu.se/en/person/MarcusNystrom/) labeled as "Manually annotated eye-tracker data".*

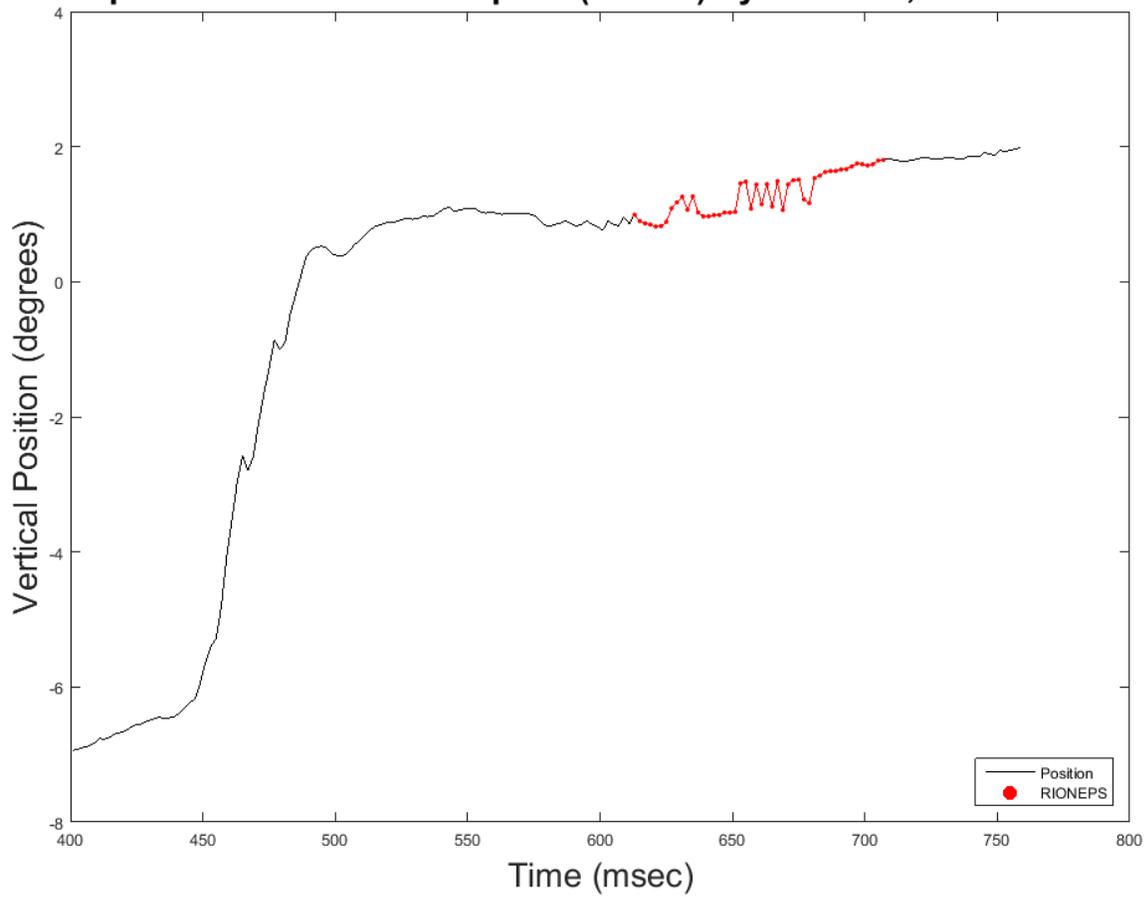

*Figure 4: The second of two examples from the same subject.  See Figure Caption for Figure 3.*

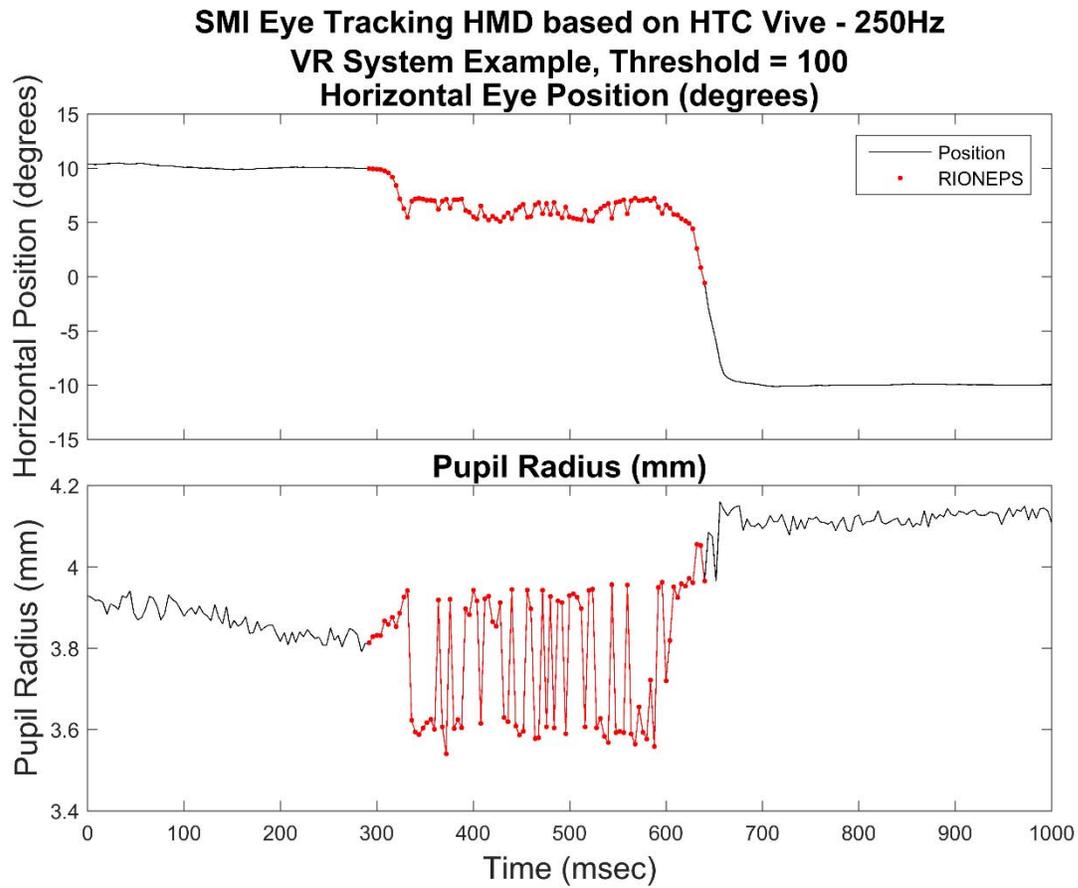

*Figure 5: These data are from a virtual reality headset fitted with an SMI Hi-Speed eye tracker operating at 250 Hz [9].*

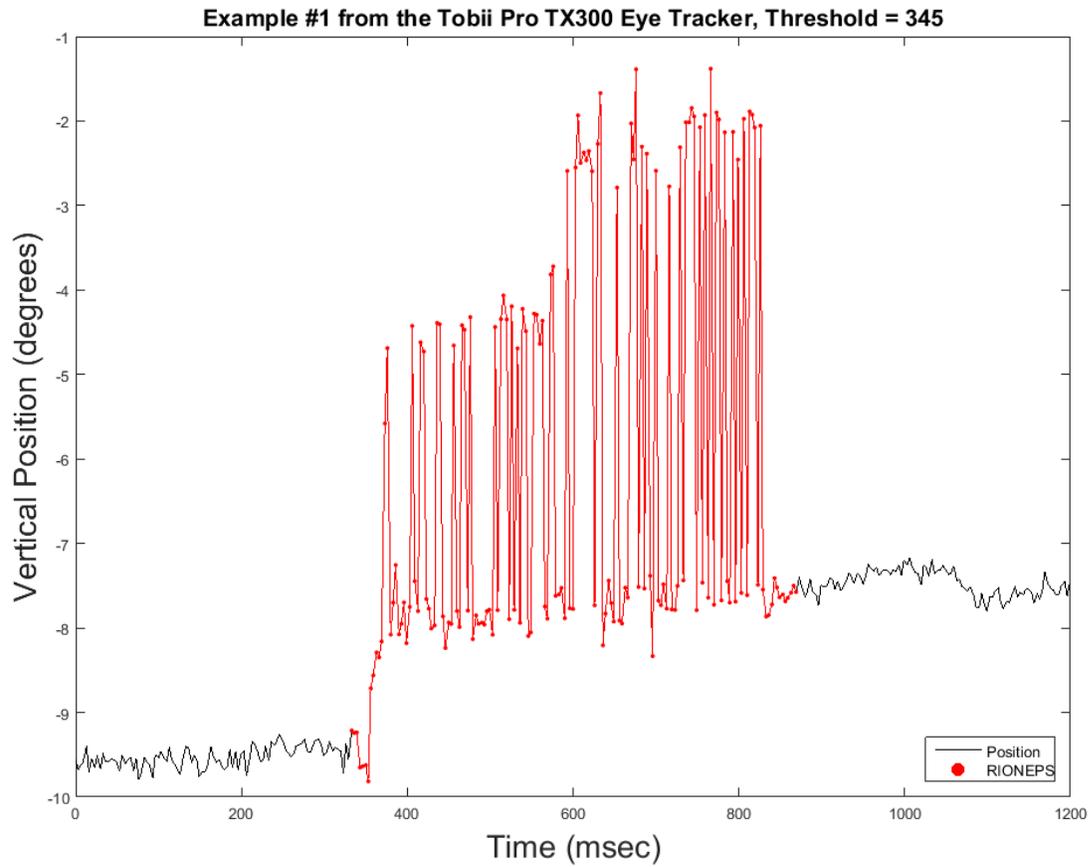

*Figure 6: Here we present the first of two examples of RIONEPS from two different subjects recorded using the Tobii Pro TX300 eye tracker [10].*

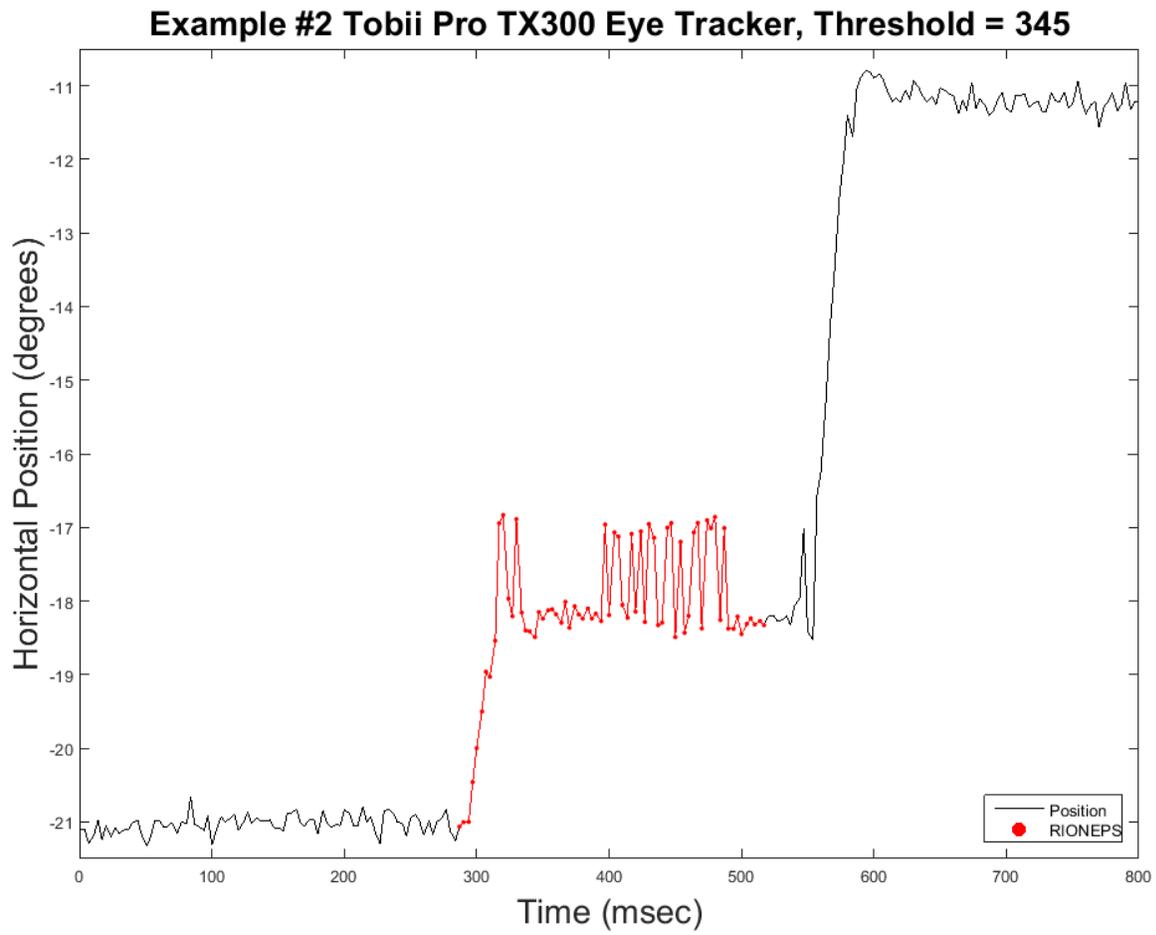

*Figure 7: The second of two examples of RIONEPS recorded using the Tobii Pro TX300 eye tracker [10].*

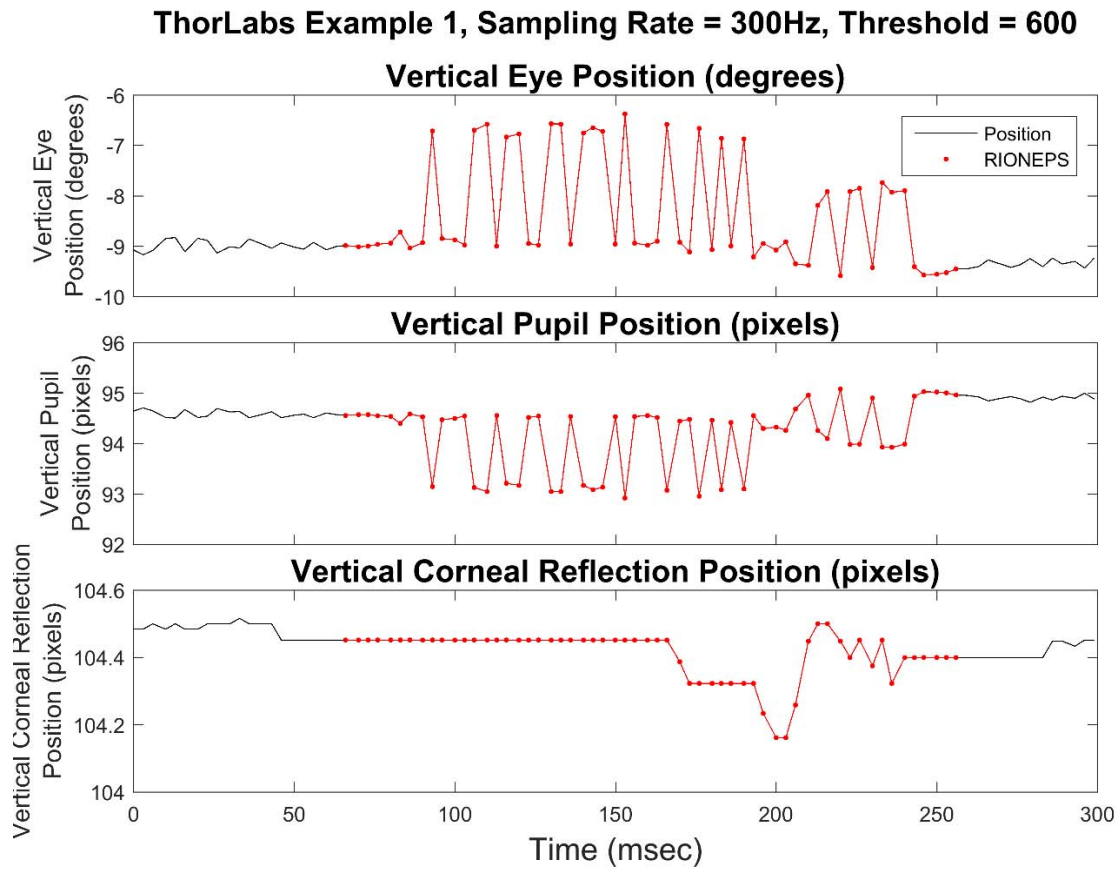

*Figure 8: These data are recorded from a custom built VOG system based on a camera from ThorLabs, at 300Hz. The custom-built system is described more fully in the text, but it does allow us to obtain instantaneous estimates of pupil position and corneal reflection position. In this case, the RIONEPS occurs in concert with changes in estimated vertical pupil position and not corneal reflection position.*

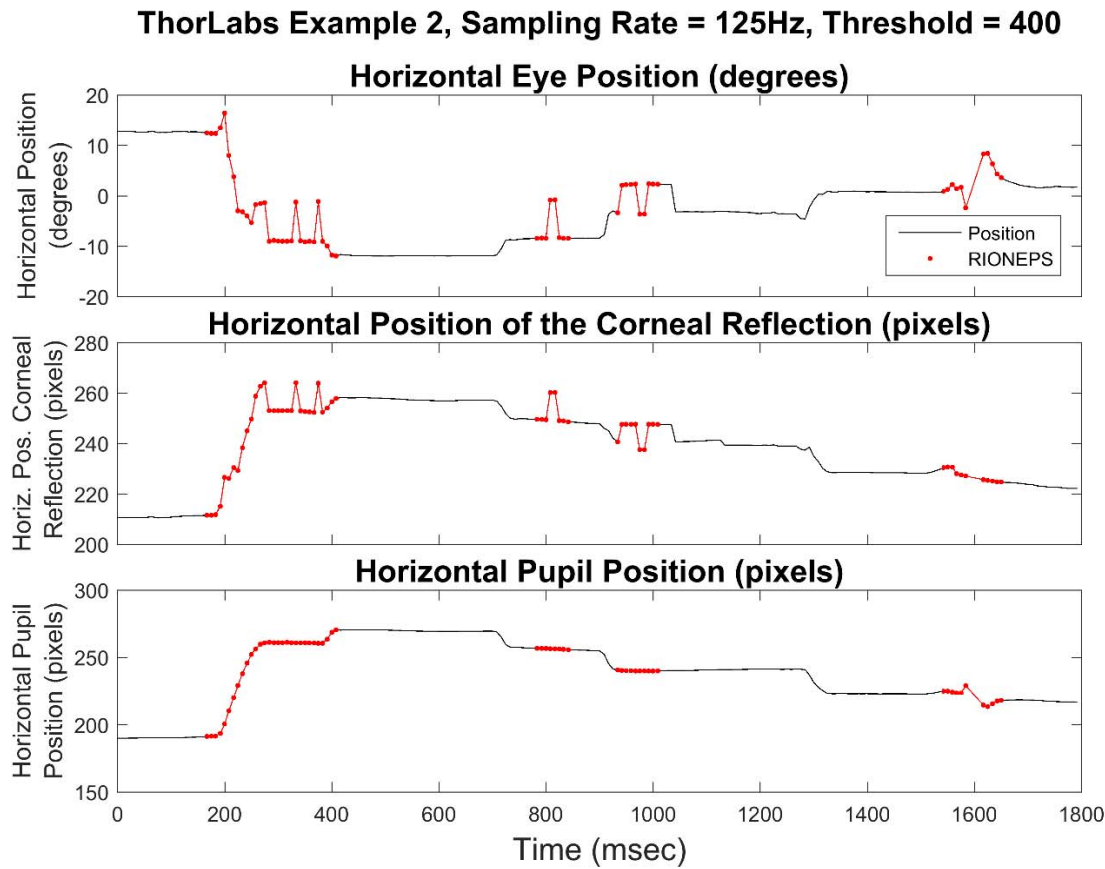

*Figure 9: These data are recorded from the same custom built VOG system mentioned in the caption of Figure 7. The data in this case were recorded at 125 Hz. Note the synchrony between the horizontal eye position trace and the horizontal position of the corneal reflection.*

**Description of the ThorLabs Custom Built System:**

To allow us to have greater feedback and control of a video-oculography setup than commercial eye trackers provide, we built a custom video-oculography setup. We refer to this setup as the ThorLabs setup, since this is the vendor that supplied the camera. This custom setup generated frames (an image of the eye), and provided data on pupil and corneal reflection position (horizontal and vertical). The hardware of the setup includes a Thorlabs (Newton, NJ) DCC1545M monochrome camera that supports various resolutions up to 1280x1024 pixels.  Lower resolutions allow a higher frame rate.  For the present report, we employed a resolution of 640 x 248 pixels at 125 frames per second (fps) and 320 x 174 pixels at 300 fps.  The camera has a Navitar (Rochester, NY) MVL7000 tele-photo lens.  A hot mirror glass (custom built) was employed to provide an image of the eye at primary position, when the camera is virtually placed in front of the eye (monocular mode).  The monitor had screen dimensions of 374 x 300 mm and a resolution of 1280 x 1024 pix, placed at the distance of 500 mm from the participant's eyes. The setup also includes a chin and forehead rest.   For software, we modified a version of an open-source eye-tracking software package known as ITU GazeTracker [11].  The modifications were designed to provide a stable high frame rate at high image resolution.  The software runs on a Windows PC.

     The first step in the algorithm is to determine where the eye image (eyebox) is within the overall image.  For this a Viola-Jones [12, 13] algorithm was used, including the use of Haar cascades [14].   Next, a pupil image intensity threshold is estimated and all pixels within the eyebox that are below the pupil image intensity threshold are indicated.   These pixels are analyzed to find a cluster – this cluster is considered to be the pupil.  The center of gravity of the pupil is taken as the pupil position.  Next, a corneal reflection threshold is estimated and all pixels within the eyebox that are brighter than this threshold are indicated.  The position of these pixels is searched to find a cluster, and this cluster is considered to be the corneal reflection.  The center of gravity of the corneal reflection is taken as the position of the corneal reflection. Next, a spatial difference vector is calculated between centers of pupil and corneal reflection.

During calibration, these spatial difference values are recorded for several calibration points. This calibration data is used to determine eye position (horizontal and vertical) during the eye-tracking task [2].

**Pseudo-Code for the Detection of RIONEPS:**

The basic idea is that, RIONEPS can be detected by calculating and thresholding the relative efficiency of the eye position signal.  Key entities in the algorithm are the total distance traveled (TDT) in a window of data, and the distance as the crow flies (DATCF). The difference between these two entities, scaled, is used to create an Inefficiency metric.  When this Inefficiency metric is greater than some threshold, the eye tracking is judged to be too inefficient for real signal and must be due to the presences of RIONEPS.  Table 1 presents the details.

--------------------------------------------------------------------------------------------------------

### Table 1: Pseudocode for our RIONEPS detection algorithm

Inputs are the position vector (PV, horizontal and vertical analyzed separately), the sample rate (SR) and the Inefficiency Threshold (IT). We begin with a logical vector called RIONEPS that is the same size as the PV, but initially all values are set to 0 (i.e., no noise):

(1) Compute Window Size (WS) = SR/20.  This will be a sliding window beginning at the first sample and ending at the last sample minus the WS. For the Eyelink-1000 the WS = 50 samples.

(2) Compute the total distance travelled (TDT) by summing the absolute value of the differences between adjacent points within the window.

(3) Compute the distance travelled in the window as the crow flies (DATCF), i.e., the absolute value of the difference between point A and point B.  In actuality, due to the potential presence of NaNs in the signal, we sum the value of the differences between adjacent points within the window and take the absolute value of this sum.

(4) Subtract DATCF from TDT, multiply this value by 1000/WS.  This is the Inefficiency metric (IM) for this window.

(5) Compute IM for all sliding windows.

(6) For each sliding window, if IM > IT, mark all points in that window as RIONEPS. (Set all points in the RIONEPS vector that correspond to this window to 1).

(7) When finished, return the RIONEPS vector, a logical vector where 0 means no noise and 1 means RIONEPS noise.

---------------------------------------------------------------------------------------------------------------

The actual Matlab (Natick, MA) code we use for this calculation is included in the Appendix.

**Discussion:**

We present a method to detect eye position signal noise that occurs in video-oculography systems, when either the detection of the pupil or the corneal reflection position is inaccurate, for whatever reason. The noise is rapidly oscillating irregularly as transient detection failures come and go. By creating an index of inefficiency (total position traveled minus distance as the crow flies), and applying a threshold to it, we can detect this noise. We see examples of this in all VOG systems tested. [1]

One drawback to our approach is that the threshold is arbitrary and will likely need to be reset for different systems. For the moment, the only way to deal with this is to find areas of RIONEPS by visual observation, and set the threshold to detect what is seen in several examples. Perhaps some automated threshold detection system could be developed in the future. Also, although we typically detect RIONEPS in offline post-processing, it is theoretically possible to detect it real time.

---

[1] Incidentally, we did not see any RIONEPS in data from a Dual Purkinje eye tracker (*15.   Engineering LfIV. Desciption of DPI database. Available from: http://live.ece.utexas.edu/research/doves/.*). This type of tracker does not use estimates of pupil or corneal reflection position (*16.   Cornsweet TN, Crane HD. Accurate two-dimensional eye tracker using first and fourth Purkinje images. J Opt Soc Am. 1973;63(8):921-8. PubMed PMID: 4722578.*).

We believe that our approach and our software will be of considerable usefulness to researchers looking to remove all noise and artifacts signal from their recorded signal.

Appendix:

Matlab Code for the Detection of RIONEPS (DetectRIONEPS.m):

```
function [RIONEPS,IM,DATCF,TDT]=DetectRIONEPS(PV,SR,IT)

%Inputs:
%   PV = Position Vector (Horizontal or Vertical)
%   SR = Sample Rate
%   IT = Inefficiency Threshold
%
%Outputs:
%   RIONEPS = Logicall vector,same length as eye_position_vector
%      0 = No Noise
%      1 = RIONEPS

RIONEPS=false(length(PV),1);

WS   =floor(SR/20); % WS = Window Size
TDT  =zeros(length(PV)-WS,1);% Total Distance Travelled
DATCF=zeros(length(PV)-WS,1);% Distance As the Crow Flies
IM   =zeros(length(PV)-WS,1);% IT Metric

FirstWindowPositionStart = 1;
LastWindowPositionStart = length(PV)-WS;
for i = FirstWindowPositionStart:LastWindowPositionStart
    %
    % Find first and last good data points
    %
    NaNsInWindow=isnan(PV(i:i+WS-1));
    NstartSkip=0;
    NaNIndex=0;
    for j = i:i+WS
        NaNIndex=NaNIndex+1;
        if NaNsInWindow(NaNIndex) == 1
```

```
            NstartSkip=NstartSkip+1;
        else
            break
        end
    end
    NendSkip=0;
    NaNIndex=WS+1;
    for j = i+WS-1:-1:i
        NaNIndex=NaNIndex-1;
        if NaNsInWindow(NaNIndex) == 1
            NendSkip=NendSkip+1;
        else
            break
        end
    end
    %
    % Compute total distance travelled (TDT) and distance as the crow flies (DATCF)
    %
    NaNIndex=0;
    CountNaNsBetweenWindowStartAndWindowEnd=0;
    for j=i+1+NstartSkip:i+WS-1-NendSkip
        NaNIndex=NaNIndex+1;
        if isnan(PV(j)) || isnan(PV(j-1))
            CountNaNsBetweenWindowStartAndWindowEnd=CountNaNsBetweenWindowStartAndWindowEnd+1;
            continue
        else
            TDT(i)=TDT(i)+abs(PV(j)-PV(j-1));
            DATCF(i)=DATCF(i)+(PV(j)-PV(j-1))
        end
    end
    DATCF(i)=abs(DATCF(i));
    %
    % Compute Inefficiency Metric (IM)
    %
    WindowSizeNoNans=WS-sum(NaNsInWindow+NstartSkip+NendSkip);
```

```
    IM(i)=(TDT(i)-DATCF(i))*1000/WindowSizeNoNans; %Inefficiency Metric (IM)
    if IM(i) < 0,IM(i) = 0;end
    if IM(i) > IT
        RIONEPS(i:i+WS-1)=1;
    end
end
return
```